\documentclass[letterpaper, 10 pt, conference]{ieeeconf}  

\IEEEoverridecommandlockouts                              

\usepackage{times}
\usepackage{graphicx}
\usepackage{subcaption}
\usepackage{caption}
\usepackage{latexsym}
\usepackage[utf8]{inputenc}
\usepackage[T1]{fontenc}

\usepackage{url}
\usepackage{breakurl}
\usepackage[breaklinks]{hyperref}

\usepackage{bm} 
\usepackage{amsmath}
\usepackage{amssymb}
\usepackage{booktabs}
\usepackage{pgf}
\usepackage{todonotes}
\usepackage{float}
\usepackage{etoolbox}
\usepackage{makecell}
\usepackage{esvect}
\usepackage{comment}

\usepackage{siunitx} 
\DeclareSIUnit[number-unit-product = {}]
\degree{deg}

\newcommand{\matr}[1]{\bm{#1}} 
\newcommand{\vect}[1]{\vec{#1}} 
\newcommand{\deriv}[1]{\dot{\vect{#1}}} 
\newcommand{\force}[1]{\vect{\tau}_{\text{#1}}} 
\newcommand{\rb}[0]{\mathit{RB}} 
\newcommand{\abs}[1]{\lvert #1 \rvert} 

\newcommand{\parabold}[1]{\textbf{#1}}

\usepackage{environ}
\makeatletter
\newsavebox{\measure@tikzpicture}
\NewEnviron{scaletikzpicturetowidth}[1]{%
  \def\tikz@width{#1}%
  \begin{lrbox}{\measure@tikzpicture}%
  \BODY
  \end{lrbox}%
  \pgfmathparse{#1/\wd\measure@tikzpicture}%
  \BODY
}
\makeatother

\usepackage{tikz}
\usetikzlibrary{matrix,shapes,arrows,positioning,chains,patterns,decorations.pathreplacing,calc}
\tikzset{
    operator/.style={
        draw,
        circle,
        thin,
        minimum height=1em,
	   inner sep=1pt
    },
    weight/.style={
        draw,
        thin,
        rounded corners,
        rectangle,
        minimum height=2em,
        minimum width=4em
    },
    value/.style={
        draw,
        thin,
        rectangle,
        minimum height=2em,
        minimum width=2.5em
    },
    gain/.style={
        regular polygon, 
        regular polygon sides=3,
        draw, 
        fill=white, 
        text width=1em,
        inner sep=1mm, 
        outer sep=0mm,
        shape border rotate=-90
    },
    concat/.style={
        draw,
        shape=circle, 
        fill=black,
        minimum height=0.5em,
	   inner sep=0pt
    },
    empty node/.style={},
    to/.style={->,>=stealth',semithick,font=\sffamily\footnotesize},
    todash/.style={->,>=stealth',dashed,semithick,font=\sffamily\footnotesize},
    every node/.style={align=center},
    diagonal fill/.style 2 args={fill=#2, path picture={
        \fill[#1, sharp corners] (path picture bounding box.south west) -|
                         (path picture bounding box.north east) -- cycle;}},
            reversed diagonal fill/.style 2 args={fill=#2, path picture={
        \fill[#1, sharp corners] (path picture bounding box.north west) |- 
                         (path picture bounding box.south east) -- cycle;}}
}

\definecolor{C1}{HTML}{ff7f0e}
\definecolor{C0}{HTML}{1f77b4}
\definecolor{RED}{HTML}{ff0000}
\definecolor{BLUE}{HTML}{0000ff}

\title{\LARGE \bf
Hybrid Physics and Deep Learning Model\\ for Interpretable Vehicle State Prediction
}

\author{Alexandra Baier$^{1}$ and Zeyd Boukhers$^{2}$ and Steffen Staab$^{1,3}$
\thanks{This work has been submitted to the IEEE for possible publication. Copyright may be transferred without notice, after which this version may no longer be accessible.}
\thanks{$^{1}$Analytic Computing, IPVS,
        University of Stuttgart, Germany
        {\tt\small {Alex.Baier,Steffen.Staab}@ipvs.uni-stuttgart.de}}%
\thanks{$^{2}$Institute for Web Science and Technologies, University of Koblenz, Germany
        {\tt\small boukhers@uni-koblenz.de}}%
\thanks{$^{3}$Web and Internet Science Research Group, University of Southampton, United Kingdom}%
}

\begin{document}

\maketitle
\thispagestyle{empty}
\pagestyle{empty}

\begin{abstract}
Physical motion models offer interpretable predictions for the motion of vehicles.
However, some model parameters, such as those related to aero- and hydrodynamics, are expensive to measure and are often only roughly approximated reducing prediction accuracy.
Recurrent neural networks achieve high prediction accuracy at low cost, as they can use cheap measurements collected during routine operation of the vehicle, but their results are hard to interpret.
To precisely predict vehicle states without expensive measurements of physical parameters,
we propose a hybrid approach combining deep learning and physical motion models including a novel two-phase training procedure.
We achieve interpretability by restricting the output range of the deep neural network as part of the hybrid model, which limits the uncertainty introduced by the neural network to a known quantity.
We have evaluated our approach for the use case of ship and quadcopter motion.
The results show that our hybrid model can improve model interpretability with no decrease in accuracy compared to existing deep learning approaches.
\end{abstract}

\section{INTRODUCTION}

Models for multi-step prediction yield a sequence of future system states given the initial system state and a sequence of control inputs. Control methods, such as model predictive control \cite{Li2020,Wu2020}, rely on the predictions of future states for computing optimal control inputs given a set of objectives like heading, speed or minimal fuel consumption. Multi-step prediction models should be accurate and, in addition, they should be interpretable such that a domain expert can understand their predictions and identify their limitations, e.g. their failure states. 
Interpretability is considered a crucial safety concern for ML systems by governmental entities, e.g. the European Commission \cite{EU2020}, or technical inspection and product certification services, e.g. the Technical Inspection Association (TÜV) \cite{Tuv2018}.
This paper develops a model for multi-step prediction, which is accurate and interpretable, to allow for future certification and usage in vehicle control.

Traditional multi-step prediction models use physical (motion) models \cite{Fossen2011,Schoukens2019}, which are commonly based on first-principles or linear regression. The parameters of these models correspond to physical parameters of the system, which makes these approaches interpretable. However, physical models suffer from the high cost and difficulty of measuring
parameters concerning, e.g., aero- or hydrodynamics. Although these costs can be reduced by applying simplifications, such as linearization of the model around a task-dependent operating point (e.g. a constant speed for manoeuvring \cite{Fossen2011}), these simplifications introduce prediction errors, which aggregate over multiple prediction steps causing the predicted states to diverge from the actual states \cite{Mohajerin2019}.

The lack of accuracy can be reduced by deep neural networks (DNN), which have shown to be capable at simulating dynamical systems over large time spans for various domains, such as unmanned aerial vehicles \cite{Chen2017,Mohajerin2019,Punjani2015}, ships \cite{Li2017b,Woo2018}, engines \cite{Schuerholz2019}, aerodynamics \cite{Li2019}, and lake temperatures \cite{Jia2019}.
DNN-based models require little domain knowledge in comparison to physical models.
In addition, they are less expensive to build, as the parameters are automatically learned from the data.
However, DNNs suffer from low interpretability, as their parameters do not directly correspond to the parameters of the system and are therefore not physically meaningful (\textbf{Challenge C1: Interpretability}). As a consequence, they may not be used without further developments in practice, as certification bodies will not licence such technologies for use in autonomous vehicles \cite{Tuv2018}, at least in the current state.

In order to benefit from both approaches, hybrid models combine physical models and DNN, while minimizing the respective disadvantages. Improved interpretability compared to deep learning is achieved by partially retaining the interpretability of the physical model via certain structures in the model architecture or by constraining the dynamics of the DNN.
Recent approaches have employed a residual hybrid architecture, in which the DNN is trained to compensate for the error residual of the physical model \cite{Chattha2019,Mohajerin2019,Woo2018}.
\cite{Chattha2019} and \cite{Woo2018} achieve a higher interpretability compared to DNN due to the combination of physical model and DNN output for the overall prediction. \cite{Mohajerin2019} feeds the output of an additional DNN as input to the physical model. This has a negative impact on the interpretability compared to \cite{Chattha2019} and \cite{Woo2018}, since the input-output relation of the physical model is obscured by the DNN.
However, the approach allows the inclusion of unknown parameters, which are identified during training of the DNN, which makes it more flexible than \cite{Chattha2019} and \cite{Woo2018}.
However, all these three approaches only consider purely linear models as physical models, which is a considerable limitation, since most vehicle dynamics are non-linear (\textbf{Challenge C2: Non-linear physical models}).

The training of hybrid models that use non-linear or inaccurate physical models for multi-step prediction is difficult (\textbf{Challenge C3: Trainability of hybrid models}).
Due to feedback of the predicted state as input to the physical model,
errors in the prediction aggregate over time, which causes divergence in the weights of the DNN.

Another limitation of the existing work is its evaluation in use cases with low environmental disturbances, such as quadcopters operated indoors or unmanned surface vehicles operated on lakes \cite{Mohajerin2019,Woo2018}.
Environmental disturbances have a large effect on vehicle dynamics due to strong coupling between states (\textbf{Challenge C4: Environmental disturbances}).

To address these challenges, we propose a novel hybrid method for multi-step prediction of vehicle states consisting of a residual hybrid architecture with a respective training procedure. Our proposed hybrid architecture combines long short-term memory networks (LSTM) and physical models similar to \cite{Chattha2019}, but with the capacity to model a large variety of physical models composed of first-principles and regression models (\textbf{C2}).
Using our two-phase training procedure, we address the existing problem of divergence during training of hybrid models in multi-step prediction, thus allowing for inaccurate and non-linear physical models as part of the hybrid model (\textbf{C3}).
Our procedure ensures convergence of the DNN by training it to correct the single-step error of the physical model and the aggregated error over multiple prediction steps.
In addition, we guarantee interpretability of our approach by applying a constraint on the output range of the LSTM (\textbf{C1}). This enables a trade-off between interpretability of the hybrid model and accuracy by varying the strictness of the constraint. Finally, we evaluate our approach on the use-case of ship motion under environmental disturbances (\textbf{C4}) and on the use-case of quadcopter motion without disturbances.
\section{RELATED WORK}\label{section:related_work}
We have identified three families of related approaches that merge knowledge of a physical system with deep learning.

\parabold{Network architectures.} 
Deep feedforward neural networks (FNN) are frequently used to approximate complex non-linear dynamics \cite{Nichiforov2017,Punjani2015}.
These FNN are known as non-linear auto-regressive exogenous (NARX) networks.
They are given a fixed window of prior states and control inputs and predict subsequent states.
Because the size of their input window is fixed, NARX networks do not cope well with long-term dynamics.

Recurrent neural networks (RNN) are a superior alternative to FNNs for simulating dynamical systems, as their recurrent structure permits them to model complex non-linear dynamics over large time spans \cite{Feng2019,Li2019,Mohajerin2018a,Woo2018}.
The hidden state of RNN can learn latent temporal dependencies, e.g. environmental effects such as waves, which are not directly measurable with sensors.
Thus, RNNs may provide more accurate state predictions than FNNs.
A network architecture that models the physical processes of the system can also improve prediction accuracy.
Sch\"urholz et al. \cite{Schuerholz2019} employs a recurrent architecture with additional forward connections between  recurrent units, which correspond to the physical information flow between components further improving the prediction accuracy.

\parabold{Physics-based loss functions.} 
Loss functions can encode known dynamics and physical parameters as regularization terms.
In \cite{Jia2019} and \cite{Muralidhar2018} physics-based restrictions motivated by the law of energy conservation are used on the loss function to simulate lake temperatures with FNNs and RNNs.
Their evaluations show that this method allows the network to converge faster than a network with unmodified loss.
Approaches dealing with object tracking  \cite{Ren2018,Stewart2017} employ loss functions for semi-supervised learning by encoding dynamics in the loss function and punishing trivial solutions.
Including knowledge about the physical system in loss functions reduces the amount of data and time required for training the DNN and results in physically plausible predictions.
However, physics-based loss functions do not improve interpretability, as they only affect the training of the neural network.

\parabold{Hybrid models.} 
Hybrid models combine different types of models, such as physical models and DNN.
In a serial architecture, approaches like \cite{Cranmer2012} and \cite{Mohajerin2018a} either give the output of a DNN as input to a physical model or vice versa.
In either variant, the output of the second model is returned as overall prediction.
The rationale underlying these architectures is that certain inputs to a physical model, such as hydrodynamic forces, are difficult to model with first-principles, but can be predicted by a DNN from control inputs and the system state.
The serial connection limits the interpretability of the hybrid model, as the input-output relation is obscured by the neural network.
On the other hand, a parallel architecture adds the output of a physical model and the output layer of a DNN into an overall prediction \cite{Cranmer2012,Woo2018}.
This approach retains the interpretability of the physical model, if the DNN contribution is sufficiently small in comparison.
Recent work uses residual architectures, which employ serial and parallel connections in joint \cite{Chattha2019,Mohajerin2018a}.
The serial connection, which gives the output of the physical model as input to the DNN, allows the network to rectify errors of the physical model and improve accuracy over multiple prediction steps. The parallel connection sums the error residual predicted by the DNN and the physical model output for the overall prediction thereby ensuring the interpretability of the prediction.
Lastly, current approaches utilize physical models, which are either relatively accurate but expensive \cite{Chattha2019} or use simplifications such as linear models \cite{Mohajerin2018a,Woo2018}.
The former type of model is expensive to develop, while the latter limits the possible parameters present in the physical model.
\section{HYBRID MODEL AND TRAINING FOR MULTISTEP PREDICTION}\label{section:model}
Our proposed approach consists of a hybrid architecture and a two-phase training procedure.
We combine a physical model with an LSTM, which predicts the error residual of the physical model. The physical model and LSTM are described in Sections~\ref{section:whitebox} and~\ref{section:blackbox}, respectively. 
The goal of the two-phase training process is to achieve convergence. 
In the first phase, we reduce the one-step prediction error by teacher-forcing \cite{Lamb2016}, i.e. at each time step, the hybrid model receives the true prior state as input. 
In the second phase, we train with multistep prediction feeding back each predicted state as input for the next prediction.
The variables used for the model definition are summarized in Table~\ref{table:notation}.
\begin{table}[t]
\setlength{\tabcolsep}{5pt}
\centering
\caption{
Summary of variables and physical parameters used in the model. 
Matrices are denoted by bold uppercase letters. 
Vectors are denoted by lowercase letters with upper arrow. 
Scalars are denoted by lowercase letters.
}
\label{table:notation}
\begin{tabular}{@{}llrl@{}}
\toprule
Variable &  \multicolumn{3}{l}{Description} \\
\midrule
$\vect{\eta}$ & \multicolumn{3}{l}{position/attitude vector in inertial frame with}\\[-0.1em]
$x$ & x-coordinate &
$y$ & y-coordinate\\[-0.1em]
$\phi$ & roll angle &
$\psi$ & yaw angle\\
\midrule
$\vect{v}$ & \multicolumn{3}{l}{velocity vector in body frame with} \\[-0.1em]
$u$ & surge velocity (x-axis) &
$w$ & sway velocity (y-axis)\\[-0.1em]
$p$ & roll rate (x-axis) &
$r$ & yaw rate (z-axis)\\
\midrule
$\matr{M}$ & \multicolumn{3}{l}{mass and inertia matrix} \\[-0.1em]
$\matr{D}(\vect{v})$ & \multicolumn{3}{l}{hydrodynamic damping matrix} \\[-0.1em]
$\matr{C}_\rb(\vect{v})$ & \multicolumn{3}{l}{rigid-body Coriolis and centripetal matrix} \\
\midrule
$\vect{g}(\vect{\eta})$ & \multicolumn{3}{l}{restoring forces} \\[-0.1em]
$\force{control}$ & \multicolumn{3}{l}{control forces (propulsion, steering)} \\[-0.1em]
$\force{env}$ & \multicolumn{3}{l}{environmental forces}\\
\midrule
$\matr{J}(\vect{\eta})$ & \multicolumn{3}{l}{rotation from body to inertial frame} \\
\midrule
$\vect{z}_t$ & \multicolumn{3}{l}{measured system state at time step $t$} \\[-0.1em]
& \multicolumn{3}{l}{$= [u, w, p, r, \phi]$}\\[-0.1em]
$\vect{c}_t$ & \multicolumn{3}{l}{control input at time step $t$}\\
\bottomrule
\end{tabular}
\end{table}
\subsection{Overall Architecture}\label{section:overall_arch}
\begin{figure}[t]
\centering
\begin{tikzpicture}[>=stealth,->,line width=0.75pt]

\matrix[matrix of nodes, column sep=0.75em, align=center, nodes={rectangle, anchor=center}] {
	\node[value] (state) {$\vect{z}_{t}$};
	& \node[concat] (cat_input) {};
	& \node[weight] (physical) {1st-Principles};
	& \node[operator] (sum_whitebox) {$+$};
	&
	&
	& \node[operator] (sum_all) {$+$};
	& \node[value] (state_est)   {$\hat{\vect{z}}_{t+1}$}; \\[1em]
	\node[value] (control) {$\vect{c}_t$};
	& 
	& \node[weight] (semiphysical) {Regression};
	&
	& \node[concat] (cat_blackbox) {};
  & \node[weight] (blackbox) {LSTM}; \\
};

\draw (control) to [out=0,in=270,loop,looseness=0.75] (cat_input);
\draw (state) to (cat_input);
\draw[to] (cat_input) -- (physical);
\draw (cat_input) to [out=0,in=180,loop,looseness=0.75]  (semiphysical);
\draw (control) to [out=270,in=270,loop,looseness=0.5] (cat_blackbox);
\draw[to] (cat_blackbox) -- (blackbox);
\draw (semiphysical) to [out=0,in=270,loop,looseness=0.75] (sum_whitebox);
\draw[to] (physical) --  (sum_whitebox);
\draw (sum_whitebox) to [out=0,in=90,loop,looseness=0.75]  (cat_blackbox);
\draw[to] (sum_whitebox) -- (sum_all);
\draw (blackbox) to [out=0,in=270,loop,looseness=0.75]  (sum_all);
\draw[to] (sum_all) -- (state_est);
\draw (blackbox) to [out=315,in=225,loop,looseness=2] (blackbox) node[xshift=0.3em,yshift=-2.6em] {$\vect{h}_{t}$};
\draw[dashed] (state_est) to [out=90, in=90, loop, looseness=0.2] (state) node[fill=white,midway,yshift=4em] {$t := t + 1$};
\end{tikzpicture}
\vspace{-1em}
\caption{
The proposed hybrid architecture consists of physical model (regression and first-principles) and LSTM.
Concatenation of vectors is represented by a black circle. The $+$-operators corresponds to an element-wise addition.
}
\label{fig:hybrid_model}
\end{figure}
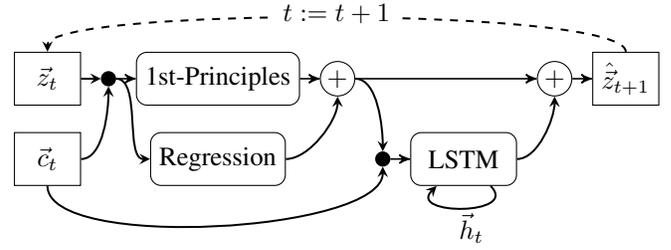

As shown in Figure~\ref{fig:hybrid_model}, the hybrid model consists of an LSTM and a physical model. The physical model can be a regression or first-principles component, as well as a combination of both. A first-principles component is a set of differential equations derived directly from Newton's second law of motion. A regression component allows learning of an interpretable model from data, which is useful estimating hydrodynamical coefficients \cite{Fossen2011}. It is common practice to combine both approaches to yield more accurate physical models.

The physical model $\vect{z}_{t+1}^{\;\text{phy}}$ computes its prediction from the control input $\vect{c}_t$ and the previous state $\vect{z}_{t}$.
The LSTM $\vect{z}^{\;\text{lstm}}_{t+1}$ predicts the error residual of the physical model  given the previous hidden state $\vect{h}_{t}$, the control input $\vect{c}_t$, and the current state prediction of the physical model $\vect{z}_{t+1}^{\;\text{phy}}$. 
The previous state $\vect{z}_{t}$ is not needed as input, since it is composed of the prior physical model output $\vect{z}_{t}^{\;\text{phy}}$, which was provided as input, and its own output $\vect{z}^{\;\text{lstm}}_{t}$.
Accordingly, this information is already encoded in the hidden state and its inclusion yielded no performance improvements in our initial experiments. 
The state prediction of the hybrid model $\hat{\vect{z}}_{t+1}$ is the sum of $\vect{z}_{t+1}^{\;\text{phy}}$ and $\vect{z}^{\;\text{lstm}}_{t+1}$.
During multi-step prediction, the auto-regressive loop, represented by the dashed arrow, feeds back this state prediction $\hat{\vect{z}}_{t+1}$ as input to the physical model.
Because of the feedback loop, the multi-step prediction error may aggregate over time.
The forward computation for our hybrid model for the time step $t+1$ is defined as follows:
\begin{align}
\begin{split}
\vect{z}^{\;\text{phy}}_{t+1} &= \text{Regression}(\vect{c}_{t}, \vect{z}_{t}) + \text{1st-Principles}(\vect{c}_{t}, \vect{z}_{t}) 
\end{split} \label{eq:physical_true}\\
\vect{h}_{t+1} &=  \text{LSTM}(\vect{c}_t, \vect{z}^{\;\text{phy}}_{t+1}, \vect{h}_{t}) \\ 
\vect{z}^{\;\text{lstm}}_{t+1} &= \matr{W}^{\text{hx}} \, \vect{h}_{t+1} \\
\begin{split}
\hat{\vect{z}}_{t+1} &= \vect{z}^{\;\text{phy}}_{t+1} + \vect{z}^{\;\text{lstm}}_{t+1} \\
\end{split}
\end{align}
\parabold{Training.} The training of the hybrid approach is divided into training of the physical model and training of the LSTM.
The regression component is fit on the residual of the first-principles model for the single-step objective:
\begin{align}
\begin{split}
\mathcal{L}^{\text{reg}}(\vect{\theta}) =& \text{MSE}(\vect{z}_{t+1} - \text{1st-Principles}(\vect{c}_{t}, \vect{z}_{t}),\\
&\phantom{\text{MSE}(}\text{Regression}(\vect{c}_t, \vect{z}_{t}; \vect{\theta}) )
\end{split}
\end{align}
If no first-principles component is used, the regression fits on the task of predicting the system state instead of the first-principles residual.

The training of the LSTM on the multi-step prediction objective, i.e. predicting the state over multiple time steps given a sequence of control inputs, is difficult, because each next-step prediction of the physical model depends on the inaccurate previous prediction. Thus, the error aggregates over several time steps causing divergence between predicted state and true state.
LSTMs are trained with backpropagation-through-time (BPTT), which is sensitive to large errors.
A combination of increasingly large errors and BPTT causes exploding gradients and prevents successful training of the network.

We solve this problem with our two-phase training procedure, which enables training hybrid models with divergent and inaccurate physical models.

\parabold{First phase.}
In the first phase, we employ teacher forcing, where the physical model receives the true state $\vect{z}_{t}$ as input at each time step $t+1$.
This mechanism enables the LSTM to learn the one-step error of the physical model, which occurs due to missing parameters or simplifications. The loss function for the first phase is defined as:
\begin{align}
\hspace*{-0.8em}\mathcal{L}^\text{lstm}(\vect{\theta}) = \sum_{t=1}^{T} \text{MSE}(\vect{z}_{t+1} - \vect{z}^{\;\text{phy}}_{t+1}, \text{LSTM}(\vect{c}_t, \vect{z}^{\;\text{phy}}_{t+1}, \vect{h}_{t}; \vect{\theta}))
\end{align}
where $T$ is the number of time steps and MSE is the mean squared error.

The first training phase is only performed to reduce the estimation error until feedback of the state prediction does not cause divergence during multi-step prediction. Therefore the first-phase loss function is only employed for a limited number of epochs.

\parabold{Second phase.}
In this phase, the model learns to predict states over multiple time steps without receiving true states as input. This is achieved by using the state prediction from the previous time step as input in the current prediction step.
The training remains effective, because the first-phase ensured that the hybrid model achieves sufficiently accurate predictions to prevent the strong divergence of DNN gradients.
The loss function for the second phase only differs from the first-phase in the physical model prediction $\vect{z}^{\;\text{phy}}_{t+1}$, as the prediction now depends on the predicted prior state $\hat{\vect{z}}_t$ and not on the true prior state:
\begin{align}
\vect{z}^{\;\text{phy}}_{t+1} &= \text{1st-Principles}(\vect{c}_t, \hat{\vect{z}}_{t}) + \text{Regression}(\vect{c}_t, \hat{\vect{z}}_{t})
\end{align}
\subsection{Physical Models}\label{section:whitebox}
We define (cf. \cite{Fossen2011,Schoukens2019}) physical models for ships combining a first-principles model with a regression model.
The simple dynamics are modelled precisely using the corresponding physical equations. The complex system dynamics are approximated using least squares regression.

\parabold{First-principles.} The models of motion are represented by differential equations, which define the equilibrium of forces resulting in the acceleration of a vehicle. 
We use the motion model for surface vessels developed by Fossen~\cite{Fossen2011}:
\begin{align}
    \deriv{\eta} &= \matr{J}(\vect{\eta}) \vect{v} \label{eq:kinematics} \\
    \deriv{v} &= \matr{M}^{{-1}} (\force{control} + \force{env} - \matr{D}(\vect{v})\vect{v}  - \matr{C}_\rb(\vect{v})\vect{v} - \vect{g}(\vect{\eta}))
	\label{eq:kinetics}
\end{align}
Equation~\ref{eq:kinematics} defines the kinematics of the vessel, as it describes the mapping of velocities in the ship's body frame to a motion in the inertial reference frame, i.e. earth. 
The ship's kinetics represented by Equation~\ref{eq:kinetics} define the relation of forces affecting the vessel and the resulting acceleration.

Developing first-principles models is challenging due to the high cost and difficulty associated with measuring their physical parameters.
Additionally, environmental variables, such as the sea state, cannot be measured during the operation of a vehicle, which makes their prediction with first-principles models infeasible.
Due to these limitations, we consider two examples of incomplete first-principles models, which entirely omit immeasurable parameters.
First, the \textbf{minimal model} (\texttt{Min}) uses only parameters that are easy to measure or approximate, such as those related to the vessel's mass, inertia and geometry. The restoring forces are linearized as $\matr{G}\vect{\eta}$ around a roll angle of $0$ yielding a good approximation with fewer parameters \cite{Fossen2011}:
\begin{align}
\deriv{v} &= \matr{M}^{{-1}} (-\matr{C}_\rb(\vect{v})\vect{v} - \matr{G} \vect{\eta}))
\end{align}
All the other parameters that are expensive to measure are set to $0$, e.g. ones concerning hydrodynamic forces.

Second, a \textbf{propulsion-based model} (\texttt{Pro}), which extends the minimal model by parameters for propulsion and steering  forces. 
This model includes $\force{control}$, which represents the control forces generated by propellers and rudders:
\begin{align}
\deriv{v} &= \matr{M}^{{-1}} (\force{control}(\vect{c}, \vect{v})-\matr{C}_\rb(\vect{v})\vect{v} - \matr{G} \vect{\eta}))
\end{align}
The model for control forces $\force{control}$ is highly dependent on the propulsion system of the modeled vessel, as it depends on the thruster configuration and the type of steering mechanism. The implementation details of a standard propeller and rudder may be found in the MSS toolbox \cite{Fossen2004}.

\parabold{Linear regression.} We employ linear regression in two models components as a coarsely approximating physical model \cite{Ljung2019}.
The \textbf{linear regression model} (\texttt{Lin}) receives  prior system state $\vect{z}_{t}$ and control input $\vect{c}_{t}$:
\begin{align}
\hat{\vect{z}}_{t+1} &= \matr{Q} \vect{c}_{t} + \matr{P} \vect{z}_{t} + \vect{b}\label{eq:linear}
\end{align}
with input matrix $\matr{Q}$, system matrix $\matr{P}$, and bias vector $\vect{b}$.

Second, we extend the linear model with a non-linear input mapping $f$ to better approximate hydrodynamic forces:
\begin{align}
\hat{\vect{z}}_{t+1} &= \matr{Q} \vect{c}_{t} +\matr{P} f(\vect{z}_{t}) + \vect{b}
\end{align}
We use the set of non-linear features developed by \cite{Blanke1993} consisting of polynomial and absolute terms, which are well-suited for modeling the damping and drag effects of a ship hull with the surrounding fluid.
%
\subsection{Recurrent Neural Network}\label{section:blackbox}
We employ a stacked LSTM including an initialization mechanism as DNN component in our hybrid model, as it is capable of memorizing long-term dynamics and of modeling dynamics at different time scales \cite{Graves2013}. The latter capability is important for modeling ships, since waves and control inputs introduce dynamics with different frequencies.

The models for multi-step prediction, such as a LSTMs, rely on an initial state to yield accurate predictions.
Thus, the initial state should encode all available information about the system and the surrounding environment, e.g. the velocity and position of a ship but also the sea state.
The state of an LSTM is its cell and hidden state. An LSTM can be initialized at $t=0$ by providing a cell and hidden state.
Mohajerin et al. \cite{Mohajerin2019} propose the initialization with an additional LSTM acting as initializer.
The initializer receives as input a window of system states and control inputs, which precede the predicted time frame.
The recurrence of the initializer enables it to extract latent dynamics from the data, which cannot be directly measured such as the sea state.
The prediction model is then initialized with the final hidden state of the initializer.
\section{EVALUATION}\label{section:evaluation}
The proposed method is evaluated extensively for modeling dynamics of a patrol ship under environmental disturbances (Section~\ref{section:expsetup} and~\ref{section:results}). To prove generalizability Section~\ref{section:quadcopter} summarizes results for quadcopter motion.
\subsection{Experiment Setup}\label{section:expsetup}
\parabold{Dataset.} The dataset is generated with a 4 degrees-of-freedom (DOF) maneuvering model of a patrol ship, which simulates the horizontal plane motion and rolling due to waves.
The original model and physical parameters are provided by Perez et al. \cite{Perez2006} and can be found in the MSS toolbox \cite{Fossen2004}.
The model is extended to include wind forces according to Isherwood \cite{Isherwood1972}, ocean waves generated with the JONSWAP spectrum \cite{Hasselmann1973}, and corresponding wave forces computed with force response amplitude operators.
Two symmetrically-placed fixed-pitch rudder propellers are implemented for control of heading and speed of the vessel.
Control inputs to the simulation model are generated via an open-control loop, which mimic a human operator.
The measurements of the vessel are sampled with a rate of \SI{1}{\hertz}.
The dataset consists of $96$ hours separated into $\SI{1}{\hour}$ simulations, where each simulation is initialized with a random sea state and control inputs.
The dataset is split into training, validation, and test set with a 60-10-30 split.

\parabold{Metrics.} The prediction performance is measured with regards to state and trajectory prediction over a time span of $\SI{15}{\minute}$ or $900$ time steps.
A window containing $\SI{1}{\minute}$ of prior control inputs and state variables is provided as initial state.
The five predicted states are the velocities for surge $u$, sway $w$, roll $p$, and yaw $r$, as well as the roll angle $\phi$.
The root mean squared error (RMSE) is averaged over all time steps for each state to measure the state prediction performance.
The trajectory RMSE measures the average distance between the predicted and true trajectory for each time step.
The trajectory error emphasizes surge, sway, and yaw, as they have the largest impact on the trajectory.
Interpretability is measured using an output range constraint enforced on the LSTM. An unconstrained LSTM results in no interpretability, while an increasingly smaller output range increases interpretability. We employ the relative threshold applied to the LSTM as measure of interpretability. The relative threshold is the ratio between the sizes of the LSTM output range and the expected output range of the physical model. The expected output range of the physical model is the $95\%$ range of its outputs computed on the training and validation set. For example, a physical model has an expected range of $[-5, 5]$ and the LSTM is constrained to a range of $[-1, 1]$, then the relative threshold is $20\%$. A lower relative threshold implies a higher degree of interpretability.

\parabold{Model selection.} Hyperparameter search is performed via grid search on the validation set.
The best model per class is chosen based on the RMSE summed over each state.
Quadratic control lag (\texttt{QLag}) is a regression model developed by \cite{Punjani2015} as best-performing physical baseline in their work.
It uses a window of prior control inputs and states, and includes control inputs as quadratic terms to model the relation between propulsion and velocity.
Stacked LSTM with an LSTM initializer (\texttt{LSTM}, \cite{Mohajerin2019}) is used as DNN baseline.
Hybrid models are trained with each potential physical model configuration using the default one-phase training and the proposed two-phase training (\{\texttt{Min}, \texttt{Pro}\}+\{\texttt{Lin}, \texttt{Pro}\}-\{\texttt{1P}, \texttt{2P}\}).
This allows us to evaluate whether our two-phase training can successfully train hybrid models with incomplete and non-linear physical models.
The models are denoted by abbreviations as defined in the previous text.
\texttt{Lin-1P} is similar to the hybrid model by \cite{Woo2018}.
\subsection{Results}\label{section:results}
\begin{table}[htbp]
\setlength{\tabcolsep}{3pt}
\centering
\caption{
Root mean squared error for each state variable and trajectory over all test samples.
\texttt{QLag} and \texttt{LSTM} are the physical and DNN baseline respectively.
Other models are hybrid models using one- or two-phase training.
The best and second-best score per column are marked in {\color{RED}red} and {\color{BLUE}blue} respectively.
}
\label{table:rmse}
\begin{tabular}{@{}lcccccc@{}}
\toprule
Model & $u$ & $w$  & $p$ & $r$  & $\phi$ & Trajectory \\[-0.1em]
& [\si{\meter\per\second}] & [\si{\meter\per\second}] & [\si{\radian\per\second}] & [\si{\radian\per\second}] & [\si{\radian}] & CI-95\% [\si{\meter}]\\
\midrule
\texttt{QLag} & 0.098 & 0.166 & \color{RED}0.0048 & 0.0058 & 0.0210 & 1147 $\pm$ \:9 \\[-0.1em]
\texttt{LSTM} & 0.085 & \color{BLUE}0.054 & 0.0056 & \color{BLUE}0.0020 & \color{RED}0.0070 & \:290 $\pm$ \:3 \\[0.2em]

\texttt{Min-1P} & 0.112 & 0.057 & 0.0057 & 0.0025 & 0.0074 & \:385 $\pm$ \:4 \\[-0.1em]
\texttt{Min-2P} & 0.144 & 0.070 & 0.0059 & 0.0023 & 0.0080 & \:401 $\pm$ \:4 \\[0.2em]

\texttt{Pro-1P} & 0.108 & 0.063 & 0.0057 & 0.0026 & 0.0079 & \:578 $\pm$ \:5 \\[-0.1em]
\texttt{Pro-2P} & 0.175 & \color{BLUE}0.054 & 0.0056 & \color{BLUE}0.0020 & 0.0075 & \:317 $\pm$ \:3 \\[0.2em]

\texttt{Lin-1P} & 0.568 & 0.645 & 0.0072 & 0.0115 & 0.0960 & 1674 $\pm$ 11 \\[-0.1em]
\texttt{Lin-2P} & 0.077 & \color{RED}0.048 & 0.0057 & \color{RED}0.0018 & \color{RED}0.0070 & \color{BLUE}\:273 $\pm$ \:3  \\[0.2em]

\texttt{Hyd-1P} &  -- & -- & -- & -- & -- & --\\[-0.1em]
\texttt{Hyd-2P} & 0.084 & 0.057 & 0.0057 & 0.0021 & \color{BLUE}0.0073 & \:315 $\pm$ \:3 \\[0.2em]

\texttt{Min+Lin-1P} & 0.828 & 0.453 & \color{BLUE}0.0052 & 0.0169 & 0.0470 & 1575 $\pm$ \:9 \\[-0.1em]
\texttt{Min+Lin-2P} & \color{BLUE}0.070 & 0.055 & 0.0060 & \color{BLUE}0.0020 & 0.0074 & \color{RED}\:269 $\pm$ \:3 \\[0.2em]

\texttt{Pro+Lin-1P} & 0.675 & 0.500 & 0.0056 & 0.0149 & 0.0591 & 1476 $\pm$ \:9 \\[-0.1em]
\texttt{Pro+Lin-2P} & 0.073 & 0.063 & 0.0062 & 0.0022 & 0.0077 & \:276 $\pm$ \:3 \\[0.2em]

\texttt{Min+Hyd-1P} & 1.667 & 0.632 & 0.0085 & 0.0143 & 0.0606 & 1798 $\pm$ 10 \\[-0.1em]
\texttt{Min+Hyd-2P} & 0.099 & 0.061 & 0.0056 & 0.0021 & 0.0077 & \:360 $\pm$ \:3 \\[0.2em]

\texttt{Pro+Hyd-1P} & 0.372 & 0.595 & 0.0106 & 0.0105 & 0.0892 & 1500 $\pm$ 10 \\[-0.1em]
\texttt{Pro+Hyd-2P} & \color{RED}0.068 & 0.063 & 0.0058 & 0.0021 & 0.0078 & \:285 $\pm$ \:3 \\
\bottomrule
\end{tabular}

\end{table}
\begin{figure*}[htbp]
\input{fig-states_time.tex}
\end{figure*}
\parabold{State prediction.} 
Table~\ref{table:rmse} shows the state prediction performance of the baselines \texttt{QLag} and \texttt{LSTM}, as well as each variation of the hybrid model. Error values across different state variables are not directly comparable, as they are not normalized and each variable has a different magnitude.
\texttt{Min-1P} is the only hybrid model without two-phase training that outperforms its counterpart with two-phase training. However, \texttt{Min-1P} performs worse overall compared to all other two-phase hybrid models.
For all other configurations, the hybrid model with one-phase training performs worse by a large factor.
For \texttt{Hyd-1P}, no model was identified to be capable of simulating the vessel without diverging completely.
These results indicate that the proposed two-step training procedure allows integration of a large range of physical models with DNN compared to the default training procedure.
\texttt{Lin-2P} achieves the lowest error for three state variables. Additionally, it achieves a trajectory error comparable to \texttt{Min+Lin-2P} as their confidence intervals overlap. We therefore consider \texttt{Lin-2P} as the best performing model in our evaluation. Overall, most hybrid models achieve performance comparable to or better than the DNN baseline \texttt{LSTM}. Hybrid models that employ \texttt{Lin} show better results on average. We hypothesize that the resulting loss function for the LSTM is less complex due to the linearity of \texttt{Lin} compared to more complex non-linear physical models and therefore easier to learn.

Figure~\ref{fig:states_over_time} visualizes a single test sample of predicted and true states over a time span of $\SI{15}{\minute}$.
Performance between \texttt{LSTM} and \texttt{Lin-2P} are similar and outperform \texttt{QLag} and \texttt{Min+Lin-1P}.
Prediction of surge speed $u$ is very accurate over \SI{15}{\minute} for \texttt{QLag}, \texttt{LSTM}, \texttt{Min+Lin-2P}. It is the easiest state to predict, since the effect of propulsion outweighs all other dynamics. However, sway $w$ and yaw rate $r$ are non-linearly coupled and strongly influenced by external forces, which makes them to difficult to predict. The impact of waves on the yaw rate $r$ is observable in the oscillations of the predicted and true state. 
\texttt{LSTM} and \texttt{Lin-2P} successfully predict the mean yaw rate, but fail to predict the wave-induced oscillations due to its randomness.
\texttt{QLag} has the best performance predicting roll rate $p$. However, the time series plot shows that the prediction is not accurate and tends towards predicting a value of $0$.
Accordingly, none of the evaluated methods are suited for use-cases with strong waves, which can be attributed to the randomness of ocean waves.

\begin{figure*}[htbp]
\input{fig-trajectory.tex}
\end{figure*}
\begin{figure*}[htbp]
\input{fig-threshold_error.tex}
\end{figure*}

\parabold{Trajectory prediction.}
The results for trajectory prediction are summarized in the last of column of Table~\ref{table:rmse}. 
The errors mirror the results for state prediction.
Hybrid models with two-phase training and \texttt{LSTM} achieve similar trajectory errors.
The hybrid models \texttt{Min-Lin-2P}, \texttt{Lin-2P} and \texttt{Pro+Lin-2P} perform significantly better than all other models.
Two-phase training provides better trajectory prediction compared to one-phase training for all cases but \texttt{Min-1P}.
Figure~\ref{fig:rmse_over_time} shows boxplots for the trajectory error per minute for the baseline model \texttt{LSTM} and the best-performing hybrid model \texttt{Lin-2P}.
Our hybrid approach achieves a lower mean and standard deviation in the trajectory error than the \texttt{LSTM}. Accordingly, the approach is more reliable for trajectory prediction over long time frames than a pure deep learning method.

An example of a trajectory prediction is visualized in Figure~\ref{fig:trajectory_pred_true}.
The trajectories are computed from the state predictions in Figure~\ref{fig:states_over_time}.
\texttt{LSTM} and \texttt{Lin-2P} predict similar trajectories, which are very close to the true trajectory.
The divergence from the true trajectory in each model occurs due to the difficulty of predicting yaw rate $r$, i.e. rotation around the z-axis, since the divergence occurs as a result of changing the heading of the vessel.
This difficulty can be observed specifically for \texttt{QLag}, as it underpredicts the yaw rate $r$, as shown in Figure~\ref{fig:qlag_states} starting at approx. \SI{300}{\second}, and consequently diverges from the true trajectory.

\parabold{Interpretability.} 
We apply a threshold to the output range of the LSTM to create an interpretable hybrid model.
Figure~\ref{fig:threshold_error} shows the RMSE of our best-performing hybrid model \texttt{Lin-2P} for surge $u$, sway $v$ and yaw rate $r$ for various relative thresholds in orange. The blue horizontal line indicates the error of the model without constraints.
The figure shows that a relative threshold of ${\sim} 15\%$ is sufficient to achieve prediction accuracy similar to the corresponding unconstrained model for all three state variables. Therefore, the loss in interpretability due to the LSTM can be limited effectively. The LSTM applies small adjustments to correct errors introduced by the physical model, which matches the training objective.

In summary, \texttt{Lin-2P} outperforms all other models including \texttt{LSTM} and the other hybrid model variants w.r.t. ship state and trajectory prediction. 
We show that \texttt{Lin-2P} can be constrained effectively to achieve model interpretability with no loss in accuracy. Accordingly, our hybrid approach enables interpretable and accurate multistep predictions in the use of case of predicting ship motion.
\subsection{Generalizability}\label{section:quadcopter}
\begin{table}[htbp]
\setlength{\tabcolsep}{2em}
\centering
\caption{
Root mean squared error for each state and trajectory over all test samples for quadcopter prediction. Only the results for hybrid models with two-phase training (\texttt{2P}) are shown.
The best and second-best scores per column are marked in {\color{RED}red} and {\color{BLUE}blue} respectively. 
}
\label{table:rmse_quadcopter}
\setlength{\tabcolsep}{0.3em}
\begin{tabular}{@{}lccccccc@{}}
\toprule
Model & $\dot{x}$ & $\dot{y}$ & $\dot{z}$& $p$& $q$& $r$& Trajectory \\[-0.1em]
& [\si{\meter\per\second}] & [\si{\meter\per\second}] & [\si{\meter\per\second}] & [\si{\radian\per\second}] & [\si{\radian\per\second}] & [\si{\radian\per\second}] & 95\% CI [\si{\meter}]\\
\midrule
\texttt{QLag} & 0.22 & 0.21 & 0.09 & 0.23 & 0.24 & 0.14 & 0.09 \\[-0.1em]
\texttt{LSTM} & \color{RED}0.08 & \color{RED}0.07 & \color{RED}0.05 & \color{RED}0.11 & \color{RED}0.11 & \color{RED}0.04 & \color{RED}0.02 \\[0.2em]

\texttt{MinQ} & 0.14 & 0.13 & 0.07 & 0.16 & 0.16 & 0.07 & \color{BLUE}0.05\\[0.2em]

\texttt{Lin} & 0.14 & 0.13 & \color{BLUE}0.06 & 0.15 & 0.15 & 0.07 & 0\color{BLUE}.05\\[-0.1em]
\texttt{Qua} & \color{BLUE}0.13 & 0.13 & \color{BLUE}0.06 & 0.15 & 0.15 & 0.07 & \color{BLUE}0.05\\[0.2em]

\texttt{MinQ+Lin} & \color{BLUE}0.13 & 0.13 & \color{BLUE}0.06 & 0.15 & 0.15 & 0.07 & \color{BLUE}0.05\\[-0.1em]
\texttt{MinQ+Qua} & \color{BLUE}0.13 & \color{BLUE}0.12 & \color{BLUE}0.06 & \color{BLUE}0.14 & \color{BLUE}0.14 & \color{BLUE}0.06 & \color{BLUE}0.05 \\
\bottomrule
\end{tabular}
\end{table}
\begin{figure*}[htbp]
\input{fig-quadcopter_threshold_error.tex}
\end{figure*}
To show generalizability of our approach, we summarize our results for the task of quadcopter motion with the dataset used in \cite{Mohajerin2019}.

\parabold{Dataset.} The dataset by \cite{Mohajerin2019} consists of multiple acrobatic maneuvers performed by an expert pilot in an indoor environment. The sampling rate is $\delta = \SI{100}{\hertz}$. Compared to the ship motion use case, measurements are more precise and frequent due to lack of environmental disturbances and use of a motion capture system.
Unlike a ship, motion of the quadcopter is mainly determined by its propulsion, as its mass and surface area are low and therefore inertial and aerodynamic forces are small. Due to the performed aerobatics, the control inputs are irregular unlike the ship use case, where control inputs were generated to simulate routine operation.
The state vector is defined as $\vect{z} = \begin{bmatrix}\dot{x} & \dot{y} & \dot{z} & p & q & r\end{bmatrix}$, which are velocities along each axis in the inertial frame and the angular rates for roll, pitch, and yaw.

\parabold{Quadcopter Models.} We perform a train-validation-test split of 60-10-30 and use grid-search to optimize hyperparameters.
As suggested by \cite{Mohajerin2019}, the evaluated prediction horizon is \SI{1}{\second} corresponding to $100$ time steps. Similar to \texttt{Min}, we develop a minimal physical model \texttt{MinQ} for quadcopters, which models rigid-body forces and gravity. We reuse the linear regression model \texttt{Lin} (Equation~\ref{eq:linear}). We introduce a quadratic regression model \texttt{Qua}, which includes the square of each control input and each state to more effectively represent forces, such as air resistance. \texttt{QLag}\cite{Punjani2015} and \texttt{LSTM}\cite{Mohajerin2019} are again used as baseline models.

\parabold{Results.} Table~\ref{table:rmse_quadcopter} shows state and trajectory prediction performance of the baseline models and each hybrid model using two-phase training for the quadcopter dataset. \texttt{LSTM} outperforms all other models in state and trajectory prediction. All hybrid models achieve similar results and outperform \texttt{QLag}. The best performing hybrid model is \texttt{MinQ+Qua-2P}. Figure~\ref{fig:quadcopter_threshold_error} shows that \texttt{MinQ+Qua-2P} constrained to a relative threshold of ${\sim} 8\%$ achieves comparable results to the unconstrained model.
The results confirm that our hybrid models are preferable w.r.t. accuracy over physical models (\texttt{QLag}). The hybrid models perform slightly worse than pure deep learning for the quadcopter motion, but allows more interpretable predictions.
\section{CONCLUSIONS}
In this paper, we developed a residual hybrid model including a two-phase training procedure. We define relative threshold as a metric for interpretability and employ constraints on the LSTM to achieve interpretable predictions.

Our approach is more flexible than existing work as it is capable of employing a large variety of physical models, such as models with missing parameters and non-linearities (\textbf{C2}: Non-linear physical models). Our results indicate that linear models are preferable as physical models w.r.t. accuracy, as they achieve the best results for ship motion prediction.
The proposed two-phase training procedure enables accurate predictions over large time spans for various physical models, while the default training approach either fails to train the network or results in worse prediction accuracy (\textbf{C3: Trainability of hybrid models}). 
The developed hybrid models outperform pure deep learning for ship motion with environmental disturbances (\textbf{C4: Environmental disturbances}) and performs only slightly worse for quadcopter motion.

The licensing of our method as part of a control system requires that the DNN contribution has to be low to ensure interpretability. Additionally, failure conditions of the DNN have to be understood to ensure safety. While we achieve the former requirement with a constraint on the DNN output range without loss of accuracy (\textbf{C1: Interpretability}), the latter one requires future research that answers what dynamics are learned by the DNN.

Additionally, our current approach only uses a simple constraint that applies to the output. Mathematical properties desirable in models for control are, for example, Lipschitz continuity and bounded-input, bounded-output stability, as they allow integration within robust model-based control frameworks. These properties require that a relation between input and output of the model is enforced. Future work will extend our hybrid approach with such model properties to integrate with model-based control frameworks.
\\\\
\noindent
\parabold{Acknowledgments.} Funded by Deutsche Forschungsgemeinschaft (DFG, German Research Foundation) under Germany's Excellence Strategy - EXC 2075 - 390740016. We acknowledge the support by the Stuttgart Center for Simulation Science (SimTech).




\bibliographystyle{IEEEtran}
\bibliography{IEEEabrv,bibliography}

\begin{thebibliography}{10}
\providecommand{\url}[1]{#1}
\csname url@rmstyle\endcsname
\providecommand{\newblock}{\relax}
\providecommand{\bibinfo}[2]{#2}
\providecommand\BIBentrySTDinterwordspacing{\spaceskip=0pt\relax}
\providecommand\BIBentryALTinterwordstretchfactor{4}
\providecommand\BIBentryALTinterwordspacing{\spaceskip=\fontdimen2\font plus
\BIBentryALTinterwordstretchfactor\fontdimen3\font minus
  \fontdimen4\font\relax}
\providecommand\BIBforeignlanguage[2]{{%
\expandafter\ifx\csname l@#1\endcsname\relax
\typeout{** WARNING: IEEEtran.bst: No hyphenation pattern has been}%
\typeout{** loaded for the language `#1'. Using the pattern for}%
\typeout{** the default language instead.}%
\else
\language=\csname l@#1\endcsname
\fi
#2}}

\bibitem{Li2020}
Z.~Li, R.~Li, and R.~Bu, ``Path following of under-actuated ships based on
  model predictive control with state observer,'' \emph{Journal of Marine
  Science and Technology}, 2020.

\bibitem{Wu2020}
Z.~Wu, D.~Rincon, and P.~Christofides, ``Process structure-based recurrent
  neural network modeling for model predictive control of nonlinear
  processes,'' \emph{Journal of Process Control}, 2020.

\bibitem{EU2020}
{European Commission}, ``Report on the safety and liability implications of
  {Artificial Intelligence}, the {Internet of Things} and robotics,'' 2020.

\bibitem{Tuv2018}
{TÜV Süd}, ``{TÜV Süd and DFKI develop ``TÜV for Artificial
  Intelligence''},'' 2018.

\bibitem{Fossen2011}
T.~Fossen, \emph{{Handbook of Marine Craft Hydrodynamics and Motion
  Control}}.\hskip 1em plus 0.5em minus 0.4em\relax Wiley, 2011.

\bibitem{Schoukens2019}
J.~Schoukens and L.~Ljung, ``{Nonlinear System Identification: {A}
  User-Oriented Roadmap},'' \emph{CoRR}, vol. abs/1902.00683, 2019.

\bibitem{Mohajerin2019}
N.~{Mohajerin} and S.~{Waslander}, ``{Multistep Prediction of Dynamic Systems
  With Recurrent Neural Networks},'' \emph{IEEE Trans. on Neural Networks and
  Learning Systems}, 2019.

\bibitem{Chen2017}
S.~Chen, Y.~Cao, Y.~Kang, R.~Zhu, and P.~Li, ``{Deep CNN Identifier for Dynamic
  Modelling of Unmanned Helicopter},'' in \emph{ICONIP}, 2017.

\bibitem{Punjani2015}
A.~{Punjani} and P.~{Abbeel}, ``Deep learning helicopter dynamics models,'' in
  \emph{ICRA}, 2015.

\bibitem{Li2017b}
G.~Li, B.~Kawan, H.~Wang, and H.~Zhang, ``Neural-network-based modelling and
  analysis for time series prediction of ship motion,'' \emph{Ship Technology
  Research}, 2017.

\bibitem{Woo2018}
J.~Woo, J.~Park, C.~Yu, and N.~Kim, ``Dynamic model identification of unmanned
  surface vehicles using deep learning network,'' \emph{Applied Ocean
  Research}, 2018.

\bibitem{Schuerholz2019}
K.~Sch{\"u}rholz, D.~Br{\"u}ckner, and D.~Abel, ``{Modelling the Exhaust Gas
  Aftertreatment System of a SI Engine Using Artificial Neural Networks},''
  \emph{Topics in Catalysis}, 2019.

\bibitem{Li2019}
K.~Li, J.~Kou, and W.~Zhang, ``{Deep neural network for unsteady aerodynamic
  and aeroelastic modeling across multiple Mach numbers},'' \emph{Nonlinear
  Dynamics}, 2019.

\bibitem{Jia2019}
X.~Jia, J.~Willard, A.~Karpatne, R.~Jordan, J.~Zwart, M.~Steinbach, and
  V.~Kumar, ``{Physics guided RNNs for modeling dynamical systems: A case study
  in simulating lake temperature profiles},'' in \emph{SIAM Int. Con. on Data
  Mining}, 2019.

\bibitem{Chattha2019}
M.~Chattha, S.~Siddiqui, M.~Malik, L.~van Elst, A.~Dengel, and S.~Ahmed,
  ``{KINN: Incorporating Expert Knowledge in Neural Networks},'' in
  \emph{AAAI-MAKE}, 2019.

\bibitem{Nichiforov2017}
C.~{Nichiforov}, I.~{Stamatescu}, I.~{Făgărăşan}, and G.~{Stamatescu},
  ``{Energy consumption forecasting using ARIMA and neural network models},''
  in \emph{5th ISEEE}, 2017.

\bibitem{Feng2019}
C.~{Feng}, L.~{Chang}, C.~{Li}, T.~{Ding}, and Z.~{Mai}, ``{Controller
  Optimization Approach Using LSTM-Based Identification Model for
  Pumped-Storage Units},'' \emph{IEEE Access}, 2019.

\bibitem{Mohajerin2018a}
N.~{Mohajerin}, M.~{Mozifian}, and S.~{Waslander}, ``{Deep Learning a Quadrotor
  Dynamic Model for Multi-Step Prediction},'' in \emph{ICRA}, 2018.

\bibitem{Muralidhar2018}
N.~{Muralidhar}, M.~{Islam}, M.~Marwah, A.~Karpatne, and N.~Ramakrishnan,
  ``{Incorporating Prior Domain Knowledge into Deep Neural Networks},'' in
  \emph{IEEE Int. Con. on Big Data}, 2018.

\bibitem{Ren2018}
H.~Ren, R.~Stewart, J.~Song, V.~Kuleshov, and S.~Ermon, ``{Learning with Weak
  Supervision from Physics and Data-Driven Constraints},'' \emph{AI Magazine},
  2018.

\bibitem{Stewart2017}
R.~Stewart and S.~Ermon, ``{Label-free Supervision of Neural Networks with
  Physics and Domain Knowledge},'' in \emph{Proc. AAAI}, 2017.

\bibitem{Cranmer2012}
A.~{Cranmer}, M.~{Shahbakhti}, and J.~{Hedrick}, ``Grey-box modeling
  architectures for rotational dynamic control in automotive engines,'' in
  \emph{ACC}, 2012.

\bibitem{Lamb2016}
A.~M. Lamb, A.~Goyal, and et~al., ``Professor forcing: A new algorithm for
  training recurrent networks,'' in \emph{NeurIPS}, 2016.

\bibitem{Fossen2004}
T.~I. Fossen and T.~Perez, ``{Marine Systems Simulator (MSS)},''
  \url{http://www.marinecontrol.org}, 2004, last accessed: 2019-09-24.

\bibitem{Ljung2019}
L.~Ljung, T.~Chen, and B.~Mu, ``A shift in paradigm for system
  identification,'' \emph{Int. Journal of Control}, 2019.

\bibitem{Blanke1993}
M.~Blanke and A.~Christensen, ``{Rudder-Roll Damping Autopilot Robustness due
  to Sway-Yaw- Roll Couplings},'' in \emph{Proc. of 10th Int. Ship Control
  Systems Symp.}, 1993.

\bibitem{Graves2013}
A.~Graves, A.~Mohamed, and G.~Hinton, ``Speech recognition with deep recurrent
  neural networks,'' in \emph{IEEE ICASSP}, 2013.

\bibitem{Perez2006}
T.~Perez, A.~Ross, and T.~Fossen, ``{A 4-DOF SIMULINK model of a coastal patrol
  vessel for manoeuvring in waves},'' in \emph{IFAC MCMC}, 2006.

\bibitem{Isherwood1972}
R.~M. Isherwood, ``Wind resistance of merchant ships,'' \emph{The Royal
  Institution of Naval Architects}, 1972.

\bibitem{Hasselmann1973}
K.~Hasselmann and D.~Olbers, ``{Measurements of wind-wave growth and swell
  decay during the Joint North Sea Wave Project},'' \emph{Erg{\"a}nzung zur
  Deut. Hydrogr. Z., Reihe}, 1973.

\end{thebibliography}

\addtolength{\textheight}{-12cm}   

\end{document}